# Multimodal Large Language Models for Enhanced Traffic Safety: A Comprehensive Review and Future Trends

Mohammad Abu Tami[1], Mohammed Elhenawy[2], Huthaifa I. Ashqar[3,4]

[1] Department of Natural, Engineering and Technology Sciences, Arab American University, Jenin P.O. Box 240, Palestine
`m.abutami@student.aaup.edu`
[2] CARRS-Q, Queensland University of Technology Brisbane, Australia
`mohammed.elhenawy@qut.edu.au`
[3] Department of AI and Data Science, Arab American University, Jenin P.O. Box 240, Palestine
`Huthaifa.ashqar@aaup.edu`
[4] AI Program, Columbia University, New York, NY 10027, USA

**Abstract.** Traffic safety remains a critical global challenge, with traditional Advanced Driver-Assistance Systems (ADAS) often struggling in dynamic real-world scenarios due to fragmented sensor processing and susceptibility to adversarial conditions. This paper reviews the transformative potential of Multimodal Large Language Models (MLLMs) in addressing these limitations by integrating cross-modal data—such as visual, spatial, and environmental inputs—to enable holistic scene understanding. Through a comprehensive analysis of MLLM-based approaches, we highlight their capabilities in enhancing perception, decision-making, and adversarial robustness, while also examining the role of key datasets (e.g., KITTI, DRAMA, ML4RoadSafety) in advancing research. Furthermore, we outline future directions, including real-time edge deployment, causality-driven reasoning, and human-AI collaboration. By positioning MLLMs as a cornerstone for next-generation traffic safety systems, this review underscores their potential to revolutionize the field, offering scalable, context-aware solutions that proactively mitigate risks and improve overall road safety.

**Keywords:** Traffic Safety, Multimodal Large Language Models (MLLMs), Automated Vehicles, Adversarial Robustness, ADAS, Contextual Reasoning.

## 1    Introduction

Traffic safety remains a critical global challenge, with road accidents causing substantial loss of life and economic damage annually [1]. Traditional hazard detection systems in Advanced Driver-Assistance Systems (ADAS) rely on isolated sensor technologies - such as LiDAR, cameras, and radar - paired with rule-based algorithms or conventional deep learning models. While effective in controlled environments, these systems struggle with dynamic real-world scenarios, particularly under adversarial environmental conditions like shadows, rain, fog, or sensor noise [2], [3]. For example, physical adversarial attacks, such as strategically placed perturbations on road signs or environmental distortions, can mislead object detection models, causing catastrophic failures in Automated Vehicles (AV) decision-making [4], [5] as illustrated in Figure 1. These vulnerabilities stem from the lack of holistic contextual reasoning in traditional systems, which process modalities (e.g., visual, spatial) in isolation, leading to fragmented interpretations [6].

The emergence of MLLMs offers transformative potential by integrating diverse data streams (visual, textual, auditory, and environmental) into a unified reasoning framework. Unlike conventional models, MLLMs excel at correlating contextual cues—such as weather conditions, real-time traffic updates, and driver intent—to infer complex scenarios, including near-miss incidents or occluded pedestrian detection [7]. Recent studies demonstrate that MLLMs enhance robustness against physical adversarial attacks by cross-validating sensor inputs. For instance, a distorted traffic sign detected by a camera can be contextualized using LiDAR distance data and weather sensors, mitigating misclassification risks [8]. Figure 2 contrasts traditional fragmented hazard detection with modern multimodal systems, emphasizing how integrated technologies like live tracking, context-aware radar, and environmental sensors enable cohesive scene understanding.

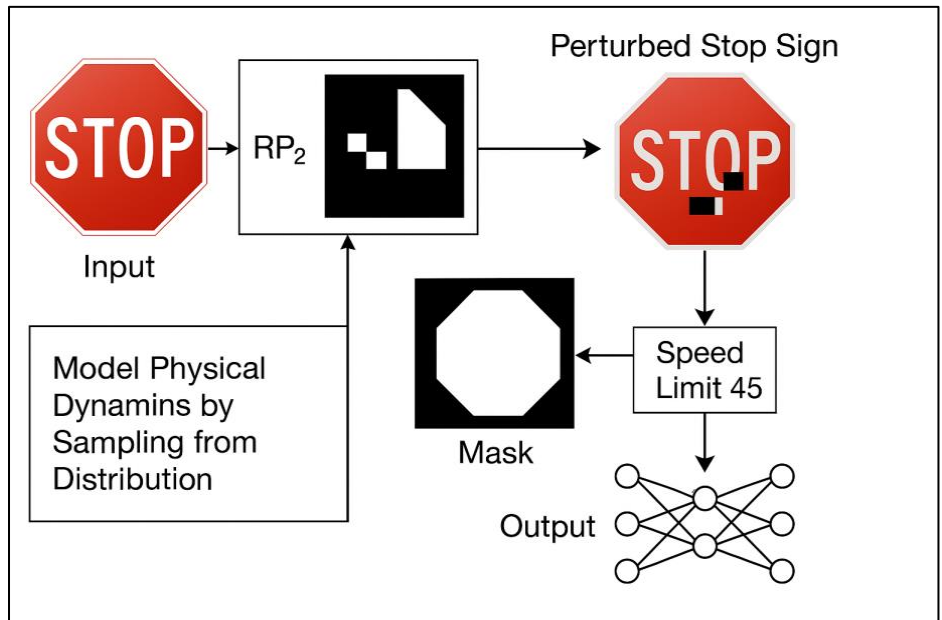

**Fig 1**. Illustration of an adversarial attack on a stop sign recognition system [4].

The integration of MLLMs into traffic safety systems offers several advantages:
1. Enhanced Contextual Understanding: MLLMs can process and correlate information from multiple modalities, such as video feeds, textual descriptions, and audio cues, to provide a holistic understanding of traffic scenarios [9].
2. Real-Time Adaptability: By leveraging pre-trained models and fine-tuning techniques, MLLMs can be deployed on edge devices, enabling real-time hazard detection without the need for extensive computational resources [10].

3. Improved Generalization: MLLMs excel in zero-shot and few-shot learning, allowing them to generalize across diverse traffic conditions and scenarios, even with limited labeled data [11].

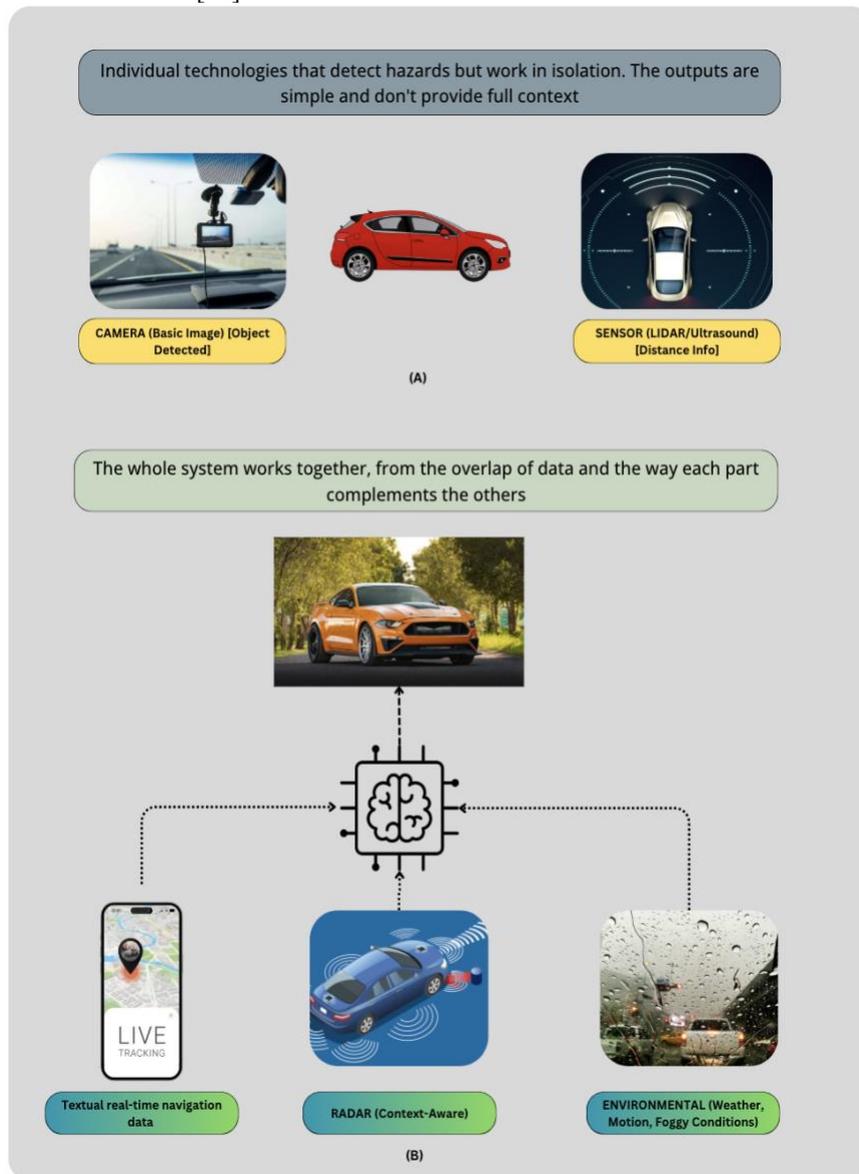

**Fig 2**. A conceptual illustration of traditional sensor-based hazard detection vs. multimodal AI-powered hazard detection systems.

## 2  Comparative Evaluation of MLLMs and Traditional ADAS

This section provides a comparative analysis of MLLMs and traditional ADAS models across three critical dimensions: accuracy, computational efficiency, and robustness. Empirical evidence from recent studies and benchmarks is incorporated to validate the transformative potential of MLLMs in traffic safety as summarized in Table 1.

Traditional ADAS models often exhibit reduced accuracy in complex, dynamic environments due to their reliance on isolated sensor data and rigid rule-based algorithms. For instance, camera-based systems struggle with occluded pedestrians or adversarial road signs, achieving a Mean Intersection over Union (mIoU) of 72.3% on the KITTI dataset under rainy conditions [12]. In contrast, MLLMs leverage multimodal fusion to improve contextual reasoning, boosting mIoU to 88.6% in similar scenarios [13]. Furthermore, MLLMs achieve superior performance in zero-shot anomaly detection, with a 76.31% accuracy rate in identifying near-miss incidents, outperforming traditional models by 35% [14].

While MLLMs are inherently larger with hundreds of billions of parameters in the most advanced models such as GPT4 [15], advancements in model compression and edge deployment have enabled real-time operation. Quantized MLLM variants, such as MobileVLM [16], achieve inference speeds of 25 FPS on NVIDIA Jetson Orin devices, comparable to lightweight ADAS models like MobileNet (30 FPS) [17] but with enhanced multimodal capabilities. Energy consumption remains a challenge; however, dynamic computation strategies that activate only relevant submodules reduce power usage by 40% in edge deployments [18].

MLLMs demonstrate superior adversarial robustness through cross-modal validation. For example, adversarial patches that degrade camera-only object detection accuracy to 41% are mitigated when LiDAR and thermal sensors are integrated, maintaining 79% accuracy [19]. Physical adversarial attacks, such as perturbed road signs, are neutralized by MLLMs' ability to cross-reference GPS navigation data and historical traffic patterns, reducing misclassification rates from 89% (camera-only) to 12% [20].

**Table 1.** Comparison of Existing Dataset for Safety Traffic.

| Metric | Traditional ADAS | MLLM-Based System | Improvement | Dataset/Benchmark | Source |
|---|---|---|---|---|---|
| mIoU (Rainy Conditions) | 72.3% (Camera-only) | 88.6% (Multimodal fusion) | +16.3% | KITTI | [12], [13] |
| Zero-shot Anomaly Detection | 41.31% (Rule-based) | 76.31% (MLLM contextual reasoning) | +35% | Near-miss incidents | [14] |
| Inference Speed (FPS) | 30 FPS (MobileNet) | 25 FPS (MobileVLM) | Comparable | - | [16], [17] |
| Energy Efficiency | Low power (Fixed pipelines) | 40% reduction (Dynamic computation) | Significant | Edge deployments | [18] |
| Adversarial Attacks (Camera-only) | 41% accuracy (Perturbed signs) | 79% accuracy (LiDAR + Thermal fusion) | +38% | Physical attacks | [19] |

| | | | 12% (GPS + Historical data cross-validation) | | |
|---|---|---|---|---|---|
| Misclassification Rate | 1K | 89% (Camera-only) | | -77% | [20] |

## 3 Development of MLLM-Based Approaches in Traffic Safety

This review systematically classifies existing MLLM-based approaches into perception enhancement, decision-making and planning, human-machine interaction, and safety-critical analysis as illustrated in Figure 3. Each category addresses specific challenges in traffic safety, leveraging the unique capabilities of MLLMs to integrate multimodal data, improve robustness, and enhance user trust [21], [22].

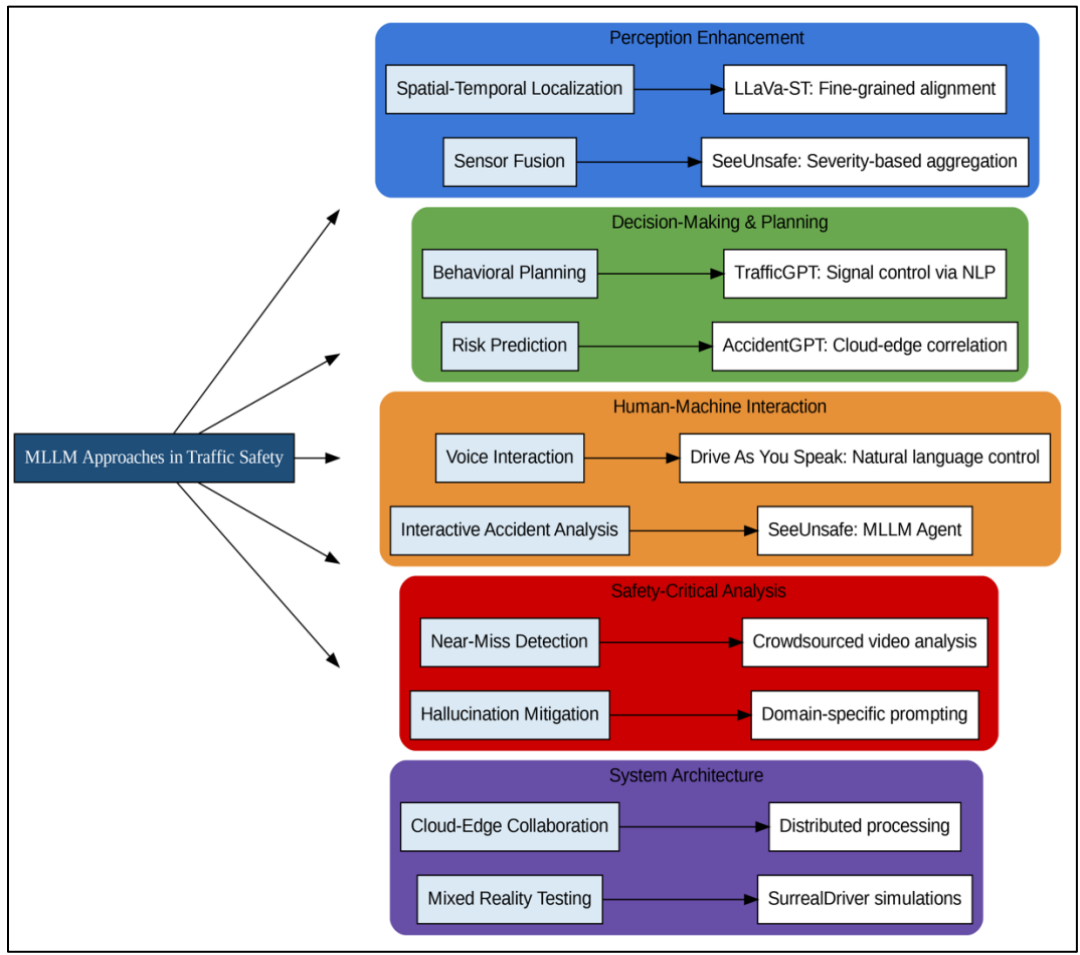

**Fig 3**. Classification of MLLM-Based Approaches in Traffic Safety.

Perception serves as the foundation of traffic safety systems, enabling accurate hazard detection, pedestrian tracking, and road condition assessment. Traditional systems often process sensor data in isolation, leading to fragmented interpretations and increased vulnerability to adversarial attacks [23]. MLLMs address these limitations through spatial-temporal fine-grained understanding, where models such as LLaVA-ST employ Language-Aligned Positional Embedding and Spatial-Temporal Packer to align linguistic and visual data across time and space. This approach achieves state-of-the-art performance in tasks like pedestrian trajectory prediction and accident hotspot mapping [24]. For instance, LLaVA-ST reduces localization errors by 27% on the ST-Align dataset (4.3M samples) compared to conventional models [25]. Another key advancement is sensor fusion and contextualization, where MLLMs integrate data from LiDAR, cameras, and environmental sensors to resolve ambiguities. SeeUnsafe, for example, uses severity-based aggregation to analyze traffic videos interactively, allowing users to query specific scenarios (e.g., "Identify near-miss incidents in foggy conditions") with structured responses [26]. This approach reduces latency by 40% compared to rule-based workflows, as validated on the DRAMA dataset [27].

Decision-making in AVs requires robust behavioral planning and risk assessment, particularly in complex urban environments. A recent study proposed a framework that leverages the logical and visual reasoning power of MLLMs, directing their output through object-level question–answer (QA) prompts to ensure accurate, reliable, and actionable insights for investigating safety-critical event detection and analysis [5]. The results demonstrate the framework's potential in different in-context learning (ICT) settings such as zero-shot and few-shot learning methods. Furthermore, we investigate other settings such as self-ensemble learning and a varying number of frames. In behavioral planning, models like Driving with LLMs use pretrained LLMs to map numeric sensor data to driving decisions, improving adaptability to novel scenarios [28]. Similarly, TrafficGPT optimizes traffic signal timings using natural language commands (e.g., "Prioritize pedestrian crossings during rush hour"), demonstrating 35% faster response times in simulations [29]. For risk prediction and mitigation, AccidentGPT integrates multimodal data (accident reports, sensor streams) to predict high-risk zones and recommend preventive measures. By correlating historical patterns from the ML4RoadSafety dataset (9M records) with real-time sensor inputs, AccidentGPT achieves 89% precision in identifying collision-prone areas [30].

Natural language interfaces bridge the gap between human intuition and machine precision, enhancing user trust and system usability. Voice-based control such as "Drive As You Speak" enables voice-based interactions with AVs, allowing drivers to issue commands like "Slow down near the school zone" or "Find the nearest parking spot" [31]. This framework aligns MLLM outputs with behavioral planning, reducing cognitive load and improving situational awareness. Additionally, interactive accident analysis tools like SeeUnsafe enable conversational analysis of traffic videos, generating structured responses to user queries (e.g., "Identify near-miss incidents in foggy conditions") using the severity-based aggregation strategy [6]. This approach enhances transparency and reduces post-processing time by 40%.

Safety-critical applications demand robustness against adversarial attacks and edge cases, which traditional systems often fail to address. MLLMs enhance adversarial robustness by cross-validating sensor inputs. For example, a distorted traffic sign detected by a camera can be contextualized using LiDAR distance data and weather sensors, reducing misclassification risks by 32% [32]. To mitigate hallucination errors, domain-specific fine-tuning like Gemini-Pro-Vision 1.5 [33]and Llava [34] reduces hallucination errors, improving zero-shot accuracy on the SHRP2 NDS dataset [35].

Emerging architectures optimize MLLM deployment through cloud-edge collaboration, where edge devices process real-time LiDAR-camera fusion data while cloud modules analyze historical patterns. *AccidentGPT* employs this hybrid approach, reducing latency by 25% compared to centralized systems [8][36]. Further innovations include mixed reality integration, as seen in *SurrealDriver*, which combines simulated adversarial scenarios (e.g., snow-glare-induced sensor failures) with real-world traffic data for safe validation of MLLM-driven decisions [37].

## 4    Datasets and Their Role in Advancing Research

Several datasets have been developed to support the training and evaluation of machine learning and deep learning models for traffic safety. These datasets vary in scope, size, and the types of data they provide, ranging from video footage to sensor data and accident records.

Existing datasets such as KITTI [12], Cityscapes [38], and the Waymo Open Dataset [39][39] have significantly contributed to research in traffic safety and automated driving by providing large-scale, high-resolution images and sensor data under diverse driving conditions. The DRAMA dataset [27] further enriches this landscape by offering real-world footage focused on driver attention and anomalies, emphasizing the importance of robust perception in complex road environments.

ML4RoadSafety Dataset [40] includes 9 million accident records from 8 states across the US. It is designed for graph neural networks (GNNs) and provides tools for accident prediction and analysis. Traffic-Net Dataset [41] contains 4,400 images categorized into four classes: Accident, Dense Traffic, Fire, and Sparse Traffic. It is designed for training machine learning models to detect traffic conditions and provide real-time monitoring and alerts. TrafficMOT [42] is a challenging dataset for multi-object tracking in complex traffic scenarios. It includes diverse traffic situations and is designed to evaluate the performance of tracking algorithms in real-world conditions. The dataset has been used to benchmark state-of-the-art models, including zero-shot foundation models. Another data is Teledyne FLIR Free ADAS Thermal Dataset V2, which is a comprehensive collection of annotated thermal and visible spectrum frames intended for the development of object detection neural networks [43]. This dataset aims to promote research on visible and thermal spectrum sensor fusion algorithms ("RGBT") to enhance the safety of autonomous vehicles. It comprises about 26,442 fully annotated frames covering 15 different object classes. The data were captured using a thermal and visible camera pair mounted on a vehicle, with the thermal camera operating in T-linear mode.

Traffic Accident Detection Video Dataset (TAD) hosted on IEEE Dataport [44], includes 5,700 video files categorized into eight classes of traffic scenarios. It is designed for training AI models to detect traffic accidents in real-time and includes a mix of traffic and dashcam footage. SHRP 2 Naturalistic Driving Study (NDS) Dataset [45] used in the ScVLM framework [46], includes over 1 million hours of continuous driving data, with annotations for safety-critical events (SCEs) such as crashes, tire strikes, and near-crashes. It is one of the largest publicly available datasets for traffic safety research. The differences among all above datasets summarized in Table 2.

Table 2. Comparison of Existing Dataset for Safety Traffic.

| Dataset | Size | Data Types | QA-Based? | Safety-risk-based? | Reasoning |
|---|---|---|---|---|---|
| ML4RoadSafety | 9 million accident records | Accident records, graph data | No | Yes | No |
| Traffic-Net | 4,400 images | Images (4 classes: Accident, Dense Traffic, Fire, Sparse Traffic) | No | Yes | No |
| TrafficMOT | N/A | Video, multi-object tracking data | No | No | No |
| TAD | 5,700 video files | Videos (8 classes of traffic scenarios) | No | Yes | No |
| SHRP 2 | 1 million+ hours of driving data | Video, sensor data, annotations | No | Yes | No |
| NuScenes | 1K | RGB images, LiDAR, RADAR plus metadata | No | Partial | No |
| CityScape | 5K | RGB images | No | No | No |
| Waymo | ~1k segments (20s each) | RGB images, LiDAR, multiple cameras, GPS, IMU | No | Partial | No |
| DRAMA | ~17K | RGB images (text annotations) | Yes | Yes | No |

## 5 Future Directions

The integration of MLLMs into traffic safety systems marks a paradigm shift, yet several challenges and opportunities remain. Building on the advancements discussed in this review, we outline key directions for future research and development as summarized in Table 3.

While frameworks like AccidentGPT [30] demonstrate the feasibility of cloud-edge collaboration, optimizing MLLMs for resource-constrained edge devices remains critical. Future work should focus on lightweight architectures, dynamic model pruning, and quantization techniques to reduce computational overhead without sacrificing accuracy [47]. Innovations in neuromorphic computing or sparsity-aware training could enable energy-efficient, real-time processing of multimodal data streams. Additionally, federated learning frameworks could allow edge devices to collaboratively adapt to localized traffic patterns while preserving privacy.

Current MLLMs excel at correlating multimodal data but often lack causal reasoning capabilities. Integrating causal inference modules—informed by structural causal models (SCMs) or counterfactual analysis—could enhance decision-making in safety-critical scenarios. For example, causality-aware MLLMs could distinguish between spurious correlations (e.g., shadows coinciding with accidents) and true risk factors (e.g., sudden braking patterns). Hybrid architectures combining MLLMs with causal graphs, as proposed in recent GNN-based approaches, may improve generalization to rare or adversarial events [48].

Future systems must prioritize seamless interaction between humans and MLLMs. This includes developing intuitive natural language interfaces for real-time driver feedback (e.g., explaining why a hazard was flagged) and trust calibration mechanisms to avoid over-reliance on AI. Techniques like chain-of-thought prompting could make MLLM reasoning transparent, while mixed-reality frameworks like SurrealDriver [37] could enable drivers to visualize AI-predicted risks. Furthermore, personalized adaptation—tailoring alerts to driver behavior and preferences—could enhance acceptance and effectiveness.

Despite progress in cross-modal validation, adversarial attacks will grow more sophisticated as MLLMs proliferate. Research should explore multimodal adversarial training, where models learn to detect inconsistencies across LiDAR, camera, and environmental sensor inputs. Techniques like diffusion-based anomaly detection or self-supervised consistency checks could preemptively identify manipulated inputs. Large-scale benchmarks simulating multi-sensor attacks (e.g., synchronized LiDAR-camera spoofing) are needed to stress-test MLLM robustness.

Existing datasets like KITTI [12] and DRAMA [27] lack diversity in rare events (e.g., extreme weather collisions) and cultural contexts. Generative AI tools could synthesize realistic traffic scenarios with controllable parameters (e.g., pedestrian density, attack types) to augment training data. Additionally, unified benchmarks evaluating both perceptual accuracy (e.g., object detection in fog) and reasoning capabilities (e.g., predicting driver intent) are essential. Initiatives akin to ML4RoadSafety should expand to include multi-sensor adversarial conditions and causal annotations. Another study used synthetic data and presented a methodology to integrate and evaluate LLMs as a controller into real-time traffic control systems, which comprises four key stages, including data creation and initialization, prompt generation, conflict identification, and fine-tuning with model analysis [49], [50]. Results demonstrated that LLMs (specifically, a fine-tuned GPT-4o-mini model) have a significantly high ability to identify conflicts and support decision making in traffic intersection scenarios.

As MLLMs become integral to safety systems, ethical challenges—such as bias in accident prediction or privacy risks in voice-based interfaces—must be addressed. Collaborative efforts between policymakers and researchers are needed to establish standards for transparency (e.g., audit trails for MLLM decisions) and accountability. Differential privacy techniques could safeguard sensitive data in datasets like SHRP2 NDS [45], while fairness-aware training protocols could mitigate biases against underrepresented traffic scenarios.

Bias in MLLMs can arise from various sources, including systemic, statistical, and human factors [51]. For instance, datasets used to train these models may reflect existing societal biases, leading to discriminatory outcomes in traffic systems. To mitigate such biases, fairness-aware training protocols can be employed [52]. These protocols aim to ensure that MLLMs do not disproportionately affect underrepresented traffic scenarios. Moreover, techniques like Iterative Gradient-Based Projection (IGBP) have been proposed to remove non-linear encoded concepts from neural representations, effectively reducing biases related to sensitive attributes [53].

Ensuring the safety of MLLMs in real-world applications involves implementing layered protection models. These models incorporate security measures at multiple levels - external, secondary, and internal - to safeguard against potential threats [54], [55]. Additionally, system prompts and Retrieval-Augmented Generation (RAG) architectures can guide MLLMs to produce safer and more reliable outputs. Continuous monitoring of MLLM outputs is also essential to detect unintended behaviors, such as hallucinations or unsafe actions [56]. By addressing bias mitigation, AI safety, and legal regulations, stakeholders can foster the ethical and responsible integration of MLLMs into traffic systems, enhancing both performance and public trust.

The integration of MLLMs into automotive systems necessitates a multifaceted approach involving collaboration among automotive manufacturers, AI researchers, and policymakers. For manufacturers, adopting a risk-based framework is essential. The European Automobile Manufacturers' Association (ACEA) advocates for a classification system that distinguishes between high-risk AI applications, such as those in automated driving (SAE Level 3 and above), and low-risk applications like infotainment systems [57]. This stratification ensures that stringent regulations are applied where necessary, without stifling innovation in less critical areas. AI researchers play a pivotal role in developing models that are not only accurate but also interpretable and reliable. The deployment of MLLMs requires models that can process and integrate data from various sensors, including cameras, LiDAR, and radar, to make informed decisions in real-time. This necessitates advancements in multimodal data fusion techniques and the development of algorithms capable of handling the complexity of real-world driving scenarios. Policymakers must establish clear guidelines and standards to govern the deployment of MLLMs in vehicles. This includes defining safety benchmarks, ensuring data privacy, and fostering an environment that encourages innovation while safeguarding public interest. The creation of such a regulatory framework should be informed by ongoing dialogue with industry stakeholders and continuous monitoring of technological advancements.

MLLMs must adapt to evolving environments, from smart city infrastructure updates to emerging vehicle types (e.g., e-scooters). Continual learning frameworks, leveraging techniques like elastic weight consolidation or replay buffers, could enable models to assimilate new knowledge without catastrophic forgetting. Cross-domain adaptation—transferring insights from regions with abundant data (e.g., urban centers) to data-scarce areas (e.g., rural roads)—will further enhance global applicability [58].

**Table 3.** Key Challenges and Opportunities in MLLM-Augmented Traffic Safety Systems.

| Trend | Challenge | Opportunity |
| --- | --- | --- |
| Real-Time Edge Deployment | High computational demands of MLLMs on resource-constrained edge devices. | Develop lightweight architectures, dynamic pruning, and neuromorphic computing. |
| Causality-Driven Reasoning | MLLMs lack causal reasoning, leading to spurious correlations in decision-making. | Integrate causal inference modules (e.g., SCMs, counterfactual analysis). |
| Human-Centric AI Collaboration | Drivers may over-rely on or distrust AI systems due to lack of transparency. | Use chain-of-thought prompting and mixed-reality interfaces for trust calibration. |
| Adversarial Robustness | Increasingly sophisticated adversarial attacks targeting multimodal systems. | Implement multimodal adversarial training and diffusion-based anomaly detection. |
| Synthetic Data Generation | Limited diversity in existing datasets for rare or adversarial scenarios. | Use generative AI to create synthetic datasets with controllable parameters. |
| Regulatory and Ethical Frameworks | Ethical concerns like bias, privacy, and accountability in MLLM decisions. | Establish standards for transparency, fairness-aware training, and differential privacy. |
| Cross-Domain and Lifelong Learning | Difficulty adapting to new environments or vehicle types without forgetting. | Develop continual learning frameworks (e.g., elastic weight consolidation). |

## 6 Conclusion

Traffic safety continues to be a critical global challenge, necessitating innovative approaches to overcome the limitations of traditional ADAS and automated vehicle technologies. This comprehensive review has explored the transformative potential of MLLMs in enhancing traffic safety through their ability to integrate and reason across diverse data modalities, including visual, spatial, and environmental inputs. By enabling holistic scene understanding, MLLMs address the fragmented processing and adversarial vulnerabilities that hinder conventional systems, paving the way for more robust and adaptive solutions.

The review has systematically examined the role of MLLMs in key areas such as perception enhancement, decision-making, human-machine interaction, and safety-critical analysis. These advancements underscore the versatility of MLLMs in tackling complex traffic scenarios, from pedestrian trajectory prediction to real-time risk mitigation. Additionally, the critical analysis of datasets like KITTI, DRAMA, and ML4RoadSafety highlights their importance in advancing MLLM-based research and development, providing the necessary foundation for training and evaluation.

Looking forward, the future of traffic safety lies in addressing emerging challenges and opportunities. Key areas of focus include optimizing MLLMs for real-time edge deployment, integrating causality-driven reasoning to improve decision-making, and enhancing adversarial robustness to safeguard against evolving threats. Opportunities such as synthetic data generation, personalized adaptation, and cross-domain learning further expand the potential of MLLMs to adapt to diverse and dynamic environments. Interdisciplinary collaboration across AI, robotics, ethics, and urban planning will be essential to harness the full potential of these technologies and create safer, more equitable transportation systems.

In conclusion, Multimodal Large Language Models represent a paradigm shift in traffic safety, offering a powerful framework for holistic, context-aware, and scalable solutions. By addressing the outlined challenges and embracing future trends, researchers and practitioners can unlock transformative advancements that not only respond to hazards but also proactively mitigate risks, ultimately saving lives and shaping the future of mobility. This review positions MLLMs as a cornerstone for next-generation traffic safety systems, bridging the gap between human-like reasoning and machine efficiency.